\documentclass[journal]{IEEEtran}

\usepackage{graphicx}
\usepackage{booktabs}
\usepackage{tabularx}
\usepackage{array}
\usepackage{ragged2e}
\usepackage{subcaption}
\usepackage{float}
\usepackage{xcolor,soul,framed} 
\usepackage{booktabs}
\usepackage{multirow}
\usepackage{subcaption}

\colorlet{shadecolor}{yellow}

\usepackage[cmex10]{amsmath}
\usepackage{array}
\usepackage{mdwmath}
\usepackage{mdwtab}
\usepackage{eqparbox}
\usepackage{url}
\usepackage[utf8]{inputenc} 
\usepackage[T1]{fontenc}    
\usepackage{hyperref}       
\usepackage{adjustbox}
\usepackage{trimclip} 
\usepackage{pgfmath}  
\usepackage{url}            
\usepackage{booktabs}       
\usepackage{amsfonts}       
\usepackage{nicefrac}       
\usepackage{microtype}      
\usepackage{xcolor}         
\usepackage{amsmath}
\usepackage{amssymb}
\usepackage{subcaption}
\usepackage{siunitx}
\usepackage{multirow}
\usepackage{wrapfig}
\usepackage{array}
\usepackage{amsthm}
\usepackage{tabularx}
\usepackage{hyperref}
\usepackage{cleveref}
\usepackage{graphicx}

\usepackage{placeins}   
\usepackage{dblfloatfix} 

\begin{document}
\bstctlcite{IEEEexample:BSTcontrol}
    \title{SPHINX: First Explain, Then Explore}
\author{
  Nguyen Do, 
  Tue M. Cao, 
  Tien Van Do, 
  Andr\'as Hajdu, 
  Tam\'as B\'erczes, 
    My T.~Thai%
  \thanks{N. Do, T. Cao, and My T. Thai are with the University of Florida, Gainesville, Florida, USA (nguyen.do@ufl.edu; caotue@ufl.edu; mythai@cise.ufl.edu).}%
  \thanks{T. V. Do is with the Budapest University of Technology and Economics, Department of Networked Systems and Services, Budapest, Hungary (do@hit.bme.hu).}%
  \thanks{A. Hajdu and T. B\'erczes are with the University of Debrecen, Debrecen, Hungary (hajdu.andras@inf.unideb.hu ; berczes.tamas@inf.unideb.hu).}%
  \thanks{Corresponding author:  mythai@cise.ufl.edu}%
  \thanks{This work has been submitted to the IEEE for possible publication. Copyright may be transferred without notice, after which this version may no longer be accessible.}
}

\markboth{IEEE TRANSACTIONS ON INTELLIGENT TRANSPORTATION SYSTEM }{Roberg \MakeLowercase{\textit{et al.}}: High-Efficiency Diode and Transistor Rectifiers}

\maketitle

\begin{abstract}

Generating adversarial driving scenarios is critical for evaluating and improving autonomous vehicle decision-making systems in simulation. Recent approaches rely primarily on the prior knowledge of Large Language Models and Vision-Language Models to generate driving scenarios procedurally. We argue that adversarial scenes should be generated based on the failure diagnosis (e.g., indecisiveness, multi-frame inconsistency) of the driving policy to specifically address the policy's weaknesses instead of relying on prior assumptions. In this paper, we propose SPHINX, a closed-loop framework for adversarial scenario synthesis guided by a simple principle: first explain, then explore. Beyond blindly exploring the scenario space, SPHINX leverages explainable artificial intelligence methods to analyze the policy, identifying key visual concepts and their influence on policy outputs, and the uncertainty of the decisions. Given the interpretable evidence extracted from the policy's own decision process, we use a vision language model to rationalize and criticize failure modes of the current policy. These critics are then used to generate targeted adversarial scenarios for policy retraining and improvement. We demonstrate that SPHINX can highlight an interpretable account of policy failures while other adversarial scene generation cannot. Across the evaluated benchmarks and test suites, SPHINX can be applied to diverse state-of-the-art autonomous vehicle architectures and yields consistent robustness improvements over existing scenario-generation methods.

\end{abstract}


\begin{IEEEkeywords}
\hl{Autonomous Driving, Modelling and Simulation, Concept-Based Explanation, Artificial Intelligence, Adversarial Testing}
\end{IEEEkeywords}

%
\IEEEpeerreviewmaketitle


\section{Introduction}

The safe deployment of Autonomous Vehicles (AVs) in real-world environments requires rigorous validation under long-tail driving scenarios, particularly safety-critical edge cases that rarely occur in standard driving datasets but can lead to catastrophic outcomes when they arise. Since collecting such rare events at scale in the real world is costly, unsafe, and inherently limited, simulation-based evaluation has become a central component of modern AV development. A key advantage of simulation is that it enables controlled generation, replay, and analysis of challenging driving episodes, making it possible to stress-test learned driving policies beyond the support of naturally observed data. Moreover, another reason that makes this simulation-driven paradigm increasingly important is because a wide range of modern AV systems often rely on learned policy models, perception-control modules, or neural driving agents that are first trained, validated, or stress-tested in simulated environments before being transferred, adapted, or evaluated in more realistic deployment settings. 

Despite recent progress, many learned driving policies \cite{bojarski2016endendlearningselfdriving, hu2022model, chitta2022transfuser, feng2026rap} still expose systematic weaknesses when facing rare, ambiguous, or adversarial traffic situations. A policy may perform well on common driving data but fail under a sudden cut-in, an occluded pedestrian, an oncoming vehicle at an intersection, or a stationary obstacle that requires early braking and lane adjustment. These failures are often not caused by missing visual information alone. A policy may observe the relevant object, but still fail because it attends to the wrong region, over-prioritizes static road cues over dynamic actors, makes uncertain control decisions, or reacts inconsistently across time. This suggests that improving AV robustness requires diverse scenarios that are not only plausible, but also targeted to the policy's underlying failure mechanisms. Luckily, recent foundation-model-based scenario generation frameworks provide a promising direction for this problem. Methods such as ChatScene~\cite{zhang2024chatscene} and LLM-Attacker~\cite{mei2025llm,peng2025ldscenellmguideddiffusioncontrollable} use Large Language Models (LLMs) or Vision-Language Models (VLMs) to synthesize semantically meaningful driving scenes, reason about multi-agent interactions, and generate adversarial or safety-critical scenarios for evaluation. In principle, retraining driving policies on such generated scenarios can improve robustness by exposing the model to challenging interactions that are rare in ordinary datasets. These methods reduce the burden of manually designing edge cases and make it possible to explore a much larger space of plausible traffic situations. 

However, diversity alone does not guarantee that the generated scenarios are useful for improving the specific policy under test. Those approaches typically generate scenes from two sources: (1) the prior knowledge of LLMs or VLMs, or (2) outcome-level feedback such as collision occurrence, adversarial reward, etc. For the first source, an LLM may know how to describe plausible traffic interactions, but its prior knowledge is not necessarily aligned with the weaknesses of the particular driving policy being evaluated. For the second source, the same outcome or reward value can be produced by many different policy behaviors. For example, two scenarios may both lead to collision, yet one may expose delayed recognition of a critical actor, while another may expose attention to irrelevant road structures, overconfident acceleration, unstable steering, or poor response to closing-speed dynamics. As a result, generated scenes may be diverse and realistic, but not necessarily aligned with what the policy needs to learn in order to improve. Retraining on such scenarios can therefore improve performance on a narrow set of generated cases without directly correcting the underlying decision weakness. This limitation motivates scenario generation that is not only diverse, but also grounded in policy-specific failure evidence.

To address this challenge, we propose \textbf{SPHINX}, a closed-loop framework for adversarial driving scenario synthesis guided by the principle: \emph{first explain, then explore}. Instead of relying only on generic LLM priors or non-identifying outcome rewards, SPHINX first analyzes the learned driving policy during simulated rollouts and extracts policy-grounded failure evidence. This evidence includes concept-level visual grounding, predictive uncertainty over control outputs, temporal control patterns, environmental metadata, and actor-level telemetry. Together, these signals characterize not only whether the policy fails, but also what information it used, how confident its decision was, how its actions evolved over time, and which interaction pattern exposed the weakness. SPHINX then aggregates these signals into temporally coherent failure windows and passes them to a VLM-based critic, which produces structured diagnoses of unsafe or hesitant behavior. An LLM-based planner converts each diagnosis into targeted scenario modifications, which are instantiated in simulation and used for policy retraining. The key distinction is that SPHINX conditions scenario generation on policy-specific failure evidence rather than scene diversity alone. For example, if a policy reacts too late to an oncoming vehicle, SPHINX generates scenarios that stress early recognition of the critical actor, closing-speed sensitivity, and decisive evasive steering. If a policy detects a stationary bus but fails to brake or adjust its lane position, SPHINX generates scenarios that emphasize obstacle interpretation, occlusion-aware reasoning, and perception-to-control alignment. Thus, the generated scenarios are designed not only to be diverse and challenging, but also to target the behavioral mechanism that the policy needs to correct. Our contributions are threefold:
\begin{itemize}
    \item We identify the semantic misalignment between foundation
    model-based scenario generation and model-specific AV
    failures, and introduce a model-grounded framework that
    conditions scenario synthesis on internal evidence rather
    than generic priors alone.

    \item We develop a unified failure analysis pipeline that combines concept-based explanations, uncertainty estimation,
    temporal aggregation, and a VLM-based multimodal critic
    to produce structured diagnostics of AV behavior.

    \item We propose a fully closed-loop system in which failureconditioned scenarios are programmatically instantiated
    in simulation and used to retrain the AV model, leading
    to improved robustness on targeted failure cases.
\end{itemize}

\section{Background and Problem Formulation}

\subsection{Autonomous Driving over Structured Scenarios}

Autonomous driving safety evaluation can be naturally formulated as a robust sequential decision-making problem over structured driving scenarios. We consider a learned ego driving policy \(\pi_\theta(u_t \mid s_t)\), parameterized by \(\theta\), where \(s_t \in \mathcal{S}\) denotes the state observed by the ego vehicle at time \(t\). The state may include ego-vehicle information, road geometry, traffic-control states, surrounding traffic participants, and their positions, velocities, and headings. The action \(u_t=(\delta_t,\alpha_t,\beta_t) \in \mathcal{U}\) denotes the control command of the ego vehicle, such as steering $\delta_t$, throttle $\alpha_t$, and braking $\beta_t$. The surrounding agents follow background behaviors denoted by \(b_t\), and the environment evolves according to
\begin{equation}
    s_{t+1} \sim P^\xi_\theta(\cdot \mid s_t, u_t, b_t),
\end{equation}
where \(\xi \in \Xi\) denotes the structured driving scenario. We call such scenarios structured because \(\xi\) specifies episode-level rollout conditions rather than an unconstrained perturbation of a single state or image. These conditions include road layout, traffic-control configurations, initial states of traffic participants, and background-agent behavior patterns over time. Therefore, \(\xi\) is constrained by traffic semantics, physical feasibility, and simulator executability. Under scenario \(\xi\), the ego policy induces a trajectory distribution \(\tau \sim P_\theta^\xi(\tau)\). The standard objective of policy learning is to maximize the expected cumulative return
\begin{equation}
    J(\theta ; \xi)
    =
    \mathbb{E}_{\tau \sim P_\theta^\xi}
    \left[
    \sum_{t=0}^T \gamma^t r(s_t,u_t)
    \right],
\end{equation}
where \(r\) is a task-dependent driving reward where its exact form may vary across autonomous driving simulators and baseline methods. In general, this reward reflects desirable driving behavior such as reaching the route goal, making safe progress, obeying traffic constraints, avoiding infractions, and producing stable control.

\subsection{Problem Formulation: Adversarial Scenario Generation}
\label{sec:problem}
Adversarial scenario generation aims to generate scenario that the policy perform poorly. We denote a scenario-level loss \(L(\theta;\xi)\), whose exact definition also depends on method being used. Throughout this formulation, larger values of \(L(\theta;\xi)\) indicate worse behavior of the ego policy under scenario \(\xi\). Therefore, adversarial scenario generation seeks scenarios that maximize this loss:
\begin{equation}
    \xi^{\star} \in \arg\max_{\xi \in \Xi} L(\theta ; \xi).
\end{equation}
A central difficulty is that the scenario space \(\Xi\) is extremely large and highly structured. It includes discrete choices, such as road topology, traffic-control states, actor types, and which background agents interact with the ego vehicle, as well as continuous variables, such as actor trajectories, speeds, and interaction timing. More importantly, \(\Xi\) is constrained by physical feasibility, traffic semantics, and simulator executability. Thus, adversarial scenario generation is not simply an unconstrained perturbation problem, but optimization over a structured and realistic scenario space. For this reason, many recent approaches replace direct optimization over \(\Xi\) with a learned scenario generator \(q_\phi(\xi)\), or more generally a conditional generator \(q_\phi(\xi \mid c)\), where \(\phi\) denotes generator parameters and \(c\) is optional context. This yields the adversarial training objective
\begin{equation}
    \max_\phi \mathbb{E}_{\xi \sim q_\phi}[L(\theta ; \xi)],
    \qquad
\end{equation}
Thus, the generator is optimized to synthesize realistic scenarios that expose weaknesses of the current policy, while the policy is retrained to reduce its loss under such generated scenarios.

\subsection{Limitations of Outcome-Driven Scenario Generation}
\label{sec:limitations}

Despite its appeal, this formulation faces a fundamental limitation. The generator is typically guided by either generic traffic priors or downstream outcome-level supervision, such as collisions, safety violations, degraded route completion, or adversarial reward. Such signals are often too coarse to identify the policy's underlying failure mechanism. Formally, the mapping \(\xi \mapsto L(\theta ; \xi)\) is generally many-to-one: distinct scenarios may induce the same loss value while exposing different weaknesses of the policy. As a result, optimizing only for high-loss scenarios does not necessarily reveal why the policy fails, what visual evidence it relies on, or whether its decisions are made under high uncertainty.

This ambiguity creates several challenges. First, the generator may exploit superficial shortcuts or simulator-specific loopholes that increase loss without revealing semantically meaningful vulnerabilities, leading to reward hacking. Second, it may collapse to a narrow family of adversarial modes that are easy to optimize but provide poor coverage of the broader failure space. Third, the supervision signal can become sparse or saturated: when the generator is weak, few scenarios induce meaningful failures, whereas when the policy is weak, many scenarios become uniformly adversarial and therefore weakly informative. Finally, because the policy and generator co-evolve during training, the induced adversarial scenario distribution is inherently non-stationary, further complicating optimization. These challenges suggest that effective scenario generation should not be guided by reward alone, but by richer signals that better capture the policy's underlying failure modes.

\section{Related Work}
\textbf{Formal Methods.}
Early efforts on scenario generation for autonomous driving were rooted in
\emph{formal specification} and \emph{falsification}. Tools such as Breach and
S-TaLiRo framed testing as the search for counterexamples against temporal or
hybrid-system specifications, while Scenic brought this line of thinking closer
to autonomous driving by introducing a probabilistic language for structured
scene and scenario specification \cite{10.1007/978-3-642-14295-6_17,10.1007/978-3-642-19835-9_21,fremont2020formal}. Subsequent work further connected formal scenario specification with simulation and real-world test execution \cite{fremont2020formal}. These methods established strong semantic structure and interpretability, but they rely heavily on manually designed specifications, constraints, and properties. As a result, they become difficult to scale when the space of interactions, agents, and environmental variations grows.

\textbf{Failure Search.}
To overcome the rigidity of purely specification-driven testing, later work
shifted toward \emph{search-based} and \emph{reinforcement-learning-based}
critical scenario discovery. Adaptive Stress Testing (AST) modeled failure
search as a sequential decision problem and used MCTS or deep RL to identify
likely failure scenarios \cite{koren2018adaptive}. This
direction was extended to more complex settings such as multi-lane traffic and
concrete critical scenario synthesis under safety-oriented objectives
\cite{trinh2024novelframeworkadaptivestress,karunakaran2022critical}.
Related frameworks such as RL-based test generation and coverage-oriented
testing further improved the ability to expose unsafe behaviors
\cite{10172814,karunakaran2022critical}. However, these
approaches remain strongly tied to reward engineering, search objectives, and
hand-crafted parameterizations. They are often effective at finding targeted
failures, but the discovered scenarios are constrained by the manually designed
search space and therefore may lack semantic richness and broader generality.

\textbf{Learned Simulation.}
A subsequent line of work addressed this limitation by moving from explicit
search over hand-designed parameters to \emph{data-driven generative traffic
simulation}. TrafficGen learned to synthesize diverse and realistic traffic
scenarios directly from driving data, while later models such as Versatile
Behavior Diffusion (VBD) and SceneDiffuser++ improved multi-agent interaction
modeling, closed-loop rollout, and large-scale generative simulation
\cite{feng2023trafficgen,tan2025scenediffuser++}. These methods shifted the focus toward learning realistic traffic dynamics and richer multi-agent interactions directly from data. However, these methods are typically designed to model realistic traffic distributions and interactions, rather than being explicitly driven by the failure signals of a specific autonomous driving system. In other words, they are often realistic, but not necessarily model-targeted.

\textbf{Language-Based Generation.}
In parallel, the emergence of large language models opened a new direction in
which \emph{natural language} became the interface for scenario authoring.
ChatScene demonstrated that LLMs can translate natural-language scenario descriptions into executable CARLA simulations through retrieval-augmented Scenic snippet assembly. \cite{zhang2024chatscene}. Later methods expanded this idea in different ways: LeGEND adopted a top-down pipeline from functional scenarios to logical and concrete scenarios, OOD scenario generation used language models to expand long-tail cases, and Txt2Sce generated standardized OpenSCENARIO programs
directly from textual reports
\cite{tang2024legend,aasi2025generating,ji2025txt2scescenariogenerationautonomous}. These methods improved semantic flexibility and reduced the burden of manual scripting, but they exposed another limitation: the gap between natural-language descriptions and executable, high-value test scenarios. ARISE explicitly addressed this problem by introducing an iterative test-and-repair loop to improve executability and correctness \cite{poddubnyy2026ariseadaptiverefinement}. Still, even with stronger executability, these pipelines are primarily guided by textual priors, retrieved knowledge, or accident reports, rather than by direct evidence of the tested model's own weaknesses.

\textbf{Adversarial Generation.}
The most recent works push this line further toward \emph{controllable} and
\emph{adversarial} closed-loop generation. LLM-Attacker uses multiple LLM-based
agents to identify attackers and iteratively generate adversarial scenarios in a
closed-loop testing framework, while LD-Scene combines LLM guidance with latent
diffusion to generate controllable adversarial safety-critical scenarios
\cite{mei2025llm,peng2025ldscenellmguideddiffusioncontrollable}. These methods
represent a clear improvement over earlier text-to-scenario pipelines because
they move beyond passive scenario authoring toward targeted stress testing.
However, they still rely substantially on attacker selection, user prompts, or
externally specified adversarial objectives. Consequently, despite being more targeted and controllable, these methods are still primarily guided by attacker selection, user prompts, or externally specified objectives, rather than by model-grounded failure evidence extracted from the system under evaluation.


\begin{figure*}[!t]
    \centering
    \hspace*{-0.05\textwidth}%
    \includegraphics[width=0.9\textwidth]{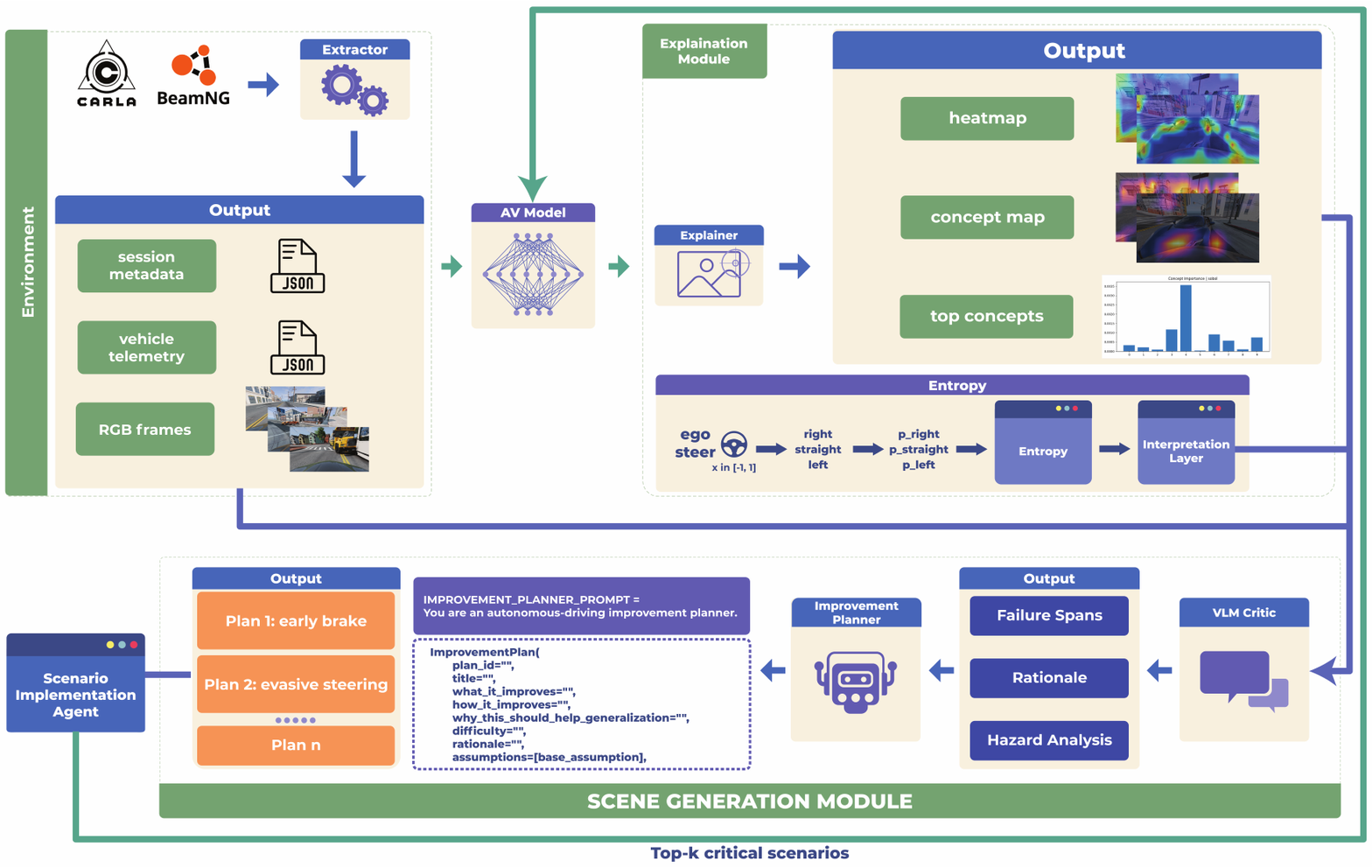}
    \caption{
Starting from simulator rollouts, a data extractor collects structured observations, which are processed by the AV model to produce trajectories and behavior signals. 
A explainer decomposes the model’s internal representation into a small set of the most important concepts, highlighting where the model focuses through concept maps and heatmaps. 
Together with control outputs and uncertainty estimates, this information is provided to a VLM critic, which performs reasoning to localize failure-relevant segments, infer underlying failure mechanisms, and assess hazard severity.
An LLM-based planner then converts the critic’s feedback into concrete scenario modifications (e.g., changing agent behavior or interaction timing) to recreate and stress the identified failure patterns. 
These scenarios are executed and validated in simulation, and the resulting data—guided by the top-$k$ key concepts driving the model’s behavior—is used to update the policy parameters $\theta$, forming a closed-loop pipeline for iterative failure discovery and policy improvement.
}
    \label{fig:overview_pipeline}
\end{figure*}

%
%

\section{The SPHINX Framework}
\label{sec:method}

We propose \textbf{SPHINX} (Fig.~\ref{fig:overview_pipeline}), a closed-loop
framework for adversarial scenario synthesis that follows the principle \emph{first explain then explore}, which we leverage our explanations of the policy's weaknesses for constructing new scenes to improve the policy model. In particular, SPHINX first explain policy's decision process on failure scene (Section \ref{sec:evidence}), highlighting the critical failure mode of the policy. Based on the evidence, we use a \textit{critic model} to rationalize the weakness of the policy (Section \ref{sec:critique}) and use the criticisms to generate new scenario to improve the model (Section \ref{sec:generation}). Lastly, the new scenes are added into dataset for closed-loop retraining (Section \ref{sec:retraining}). This directly targets the limitation discussed in Section~\ref{sec:limitations} where
two scenarios with identical loss can expose different weaknesses, our SPHINX conditions on the weakness of the decision from the policy rather than on the outcome loss alone.


\subsection{Policy-Grounded Failure Evidence Extraction}
\label{sec:evidence}

\paragraph{Goal}
Given a rollout of length $T$ produced by the current policy $\pi_\theta$, this stage
summarizes every frame by an evidence descriptor that characterizes \textit{why} the policy may be vulnerable to crashing. These evidence will be used to generate critics on the model's failures and new test scenario to improve model's policy. We characterize each frame along four complementary, policy-grounded axes:
\begin{enumerate}
  \item \emph{Where} the decision is visually grounded --- the attribution map $\bar{M}_t$ (Equation \ref{eq:attribution}).
  \item \emph{Whether} that grounding lands on task-relevant content --- the mismatch $\Gamma_t$ (Equation \ref{eq:mismatch}).
  \item \emph{How confident} the control decision is --- the predictive uncertainty $U_t$ (Equation \ref{eq:uncertainty}).
  \item \emph{How stable} the control is over time --- the temporal instability $\Delta_t$ (Equation \ref{eq:instability}).
\end{enumerate}
Together with the raw control $u_t$ and the structured scene context $g_t$ (retained for
downstream critique), these define the per-frame \emph{evidence descriptor}
\begin{equation}
  z_t \;=\; \left(\,\bar{M}_t,\;\Gamma_t,\;U_t,\;\Delta_t,\;u_t,\;g_t,\;\rho_t\,\right),
  \label{eq:descriptor}
\end{equation}
where $\rho_t$ is a scalar \emph{failure-salience} score that fuses the three
failure-indicative quantities $(\Gamma_t,U_t,\Delta_t)$ and is used to prioritize the focus of the critics on failure cases. The stage output is the evidence stream
$\mathcal{E}_{1:T}=\{z_t\}_{t=1}^{T}$. We define each component below.


\paragraph{Concept-level visual grounding $\bar{M}_t$}
We aim to understand the decision making process of the policy model $\pi_\theta(u_t\mid s_t)$ by breaking down its hidden representation $h_t$ at time $t$ into interpretable concepts. The internal concepts show the model considerations in making the final output (i.e. pavement, in-comming cars, etc.). We extract Top-K most important concepts ranked by explainer method (such as \cite{fel2023craft3}) for the decision at time $t$:  $\mathcal{K}_t=\operatorname{TopK}\left(w_{t,1},\dots,w_{t,K}\right)$. Each concept has activation $\alpha_{t,k}\in\mathbb{R}$ measuring the concept strength, and an attribution map $M_{t,k}\in[0,1]^{H\times W}$ localizing the regions of the image associated with concept $k$. We convert activations into importance weights via a temperature $\lambda$ softmax over absolute activations and aggregate the dominant concepts:
\[
w_{t,k}=\frac{\exp\!\left(\lambda|\alpha_{t,k}|\right)}{\sum_{j=1}^{K}\exp\!\left(\lambda|\alpha_{t,j}|\right)}, \qquad \lambda>0
\]

\begin{equation}
\bar{w}_{t,k}=w_{t,k}/\sum_{j\in\mathcal{K}_t}w_{t,j},
\qquad
\bar{M}_t=\sum_{k\in\mathcal{K}_t}\bar{w}_{t,k}\,M_{t,k},
  \label{eq:attribution}
\end{equation}
The map $\bar{M}_t\in[0,1]^{H\times W}$ summarizes
the visual evidence on which the policy's control at time $t$ is grounded. We use method in \cite{fel2023craft3} throughout this paper due to its fast training time, however, more advanced concept explainer can be used.

\paragraph{Semantic grounding mismatch $\Gamma_t$}
Based on the visual grounding $\bar{M}_t$, the model may focus on irrelevant regions in the image that are not important for driving (i.e. the sky, buildings). When task-relevant regions 
$\mathcal{R}_t\subseteq[H]\times[W]$ are available (lane boundaries, drivable area,
relevant vehicles, etc.), we measure the fraction
of attribution mass that falls \emph{outside} $\mathcal{R}_t$:
\begin{equation}
  \Gamma_t \;=\; 1-\frac{\langle \bar{M}_t,\mathbf{1}_{\mathcal{R}_t}\rangle}
                        {\|\bar{M}_t\|_1+\varepsilon}\;\in\;[0,1],
  \qquad \varepsilon>0,
  \label{eq:mismatch}
\end{equation}
where $\mathbf{1}_{\mathcal{R}_t}\in\{0,1\}^{H\times W}$ is the indicator function and $\langle A,B\rangle=\sum_{p,q}A_{pq}B_{pq}$ and $\|A\|_1=\sum_{p,q}|A_{pq}|$.
Larger $\Gamma_t$ indicates weaker alignment between the policy's evidence and
task-relevant content. When such masks are unavailable, this term is omitted.

\paragraph{Predictive uncertainty $U_t$}
Well-placed visual grounding does not imply a confident and accurate decision. We observe that the model indecisiveness often lead to failure. For example, the model was uncertain in choosing steering direction and decided to steer to the left by a small amount at each frame, ultimately did not steer enough to dodge the in-coming car. Therefore, to  quantify the uncertainty, we discretize each control channel
$m\in\{\delta,\alpha,\beta\}$ into $N_m$ bins and estimate channel-wise predictive
distributions $p_\theta^{m}(\,\cdot\mid s_t\,)$ over those bins using $S$ stochastic forward
passes (e.g., MC-dropout, input perturbation, or an ensemble),
$\hat{p}_\theta^{m}(i\mid s_t)=\tfrac{1}{S}\sum_{s=1}^{S}\mathbf{1}\!\left[u^{(s)}_{t,m}\in b_i^{m}\right]$.
The per-channel entropy (in nats) and the aggregate uncertainty are
\begin{equation}
  H_t^{m}=-\!\sum_{i=1}^{N_m}\hat{p}_\theta^{m}(i\mid s_t)\,
            \log \hat{p}_\theta^{m}(i\mid s_t),
  \: \:
  U_t=\!\!\sum_{m\in\{\delta,\alpha,\beta\}}\!\!\omega_m H_t^{m},
  \label{eq:uncertainty}
\end{equation}
with channel weights $\omega_\delta+\omega_\alpha+\omega_\beta=1$. Large $U_t$ indicates an
uncertain control decision at time $t$.

\paragraph{Temporal control instability $\Delta_t$}
Driving failures are typically temporal: a policy fails not from one wrong frame but from
control that evolves inconsistently. In particular, the model may decide to steer left at frame $t$ while change to steer right at the next frame, ultimately fail to avoid the obstacle in the front. We therefore quantify local instability by a finite difference of
the control sequence:
\begin{equation}
  \Delta_t=\|u_t-u_{t-1}\|_2
  \qquad\text{or}\qquad
  \Delta_t=\|u_t-2u_{t-1}+u_{t-2}\|_2,
  \label{eq:instability}
\end{equation}
using the first-order form by default and the second-order form when stronger sensitivity to
oscillation or delayed correction is required. Large $\Delta_t$ flags abrupt control changes
consistent with hesitation, oscillation, or late reaction.

\paragraph{Frame-level failure salience $\rho_t$}
Finally, we fuse the three failure-indicative quantities into a single salience score:
\begin{equation}
  \rho_t=\lambda_g\,\Gamma_t+\lambda_u\,U_t+\lambda_\Delta\,\Delta_t,
  \qquad \lambda_g,\lambda_u,\lambda_\Delta\ge 0.
  \label{eq:salience}
\end{equation}
Its role is to rank and grouping the frames into windows for the critic model in the next
Stage~\ref{sec:critique}.

\subsection{Multimodal Temporal Critique}
\label{sec:critique}

The stream $\mathcal{E}_{1:T}$ localizes vulnerability at individual frames, but unsafe
behavior emerges over time. This stage groups high-salience frames into temporal windows (Section \ref{sec:anchor}) and
asks a multimodal critic to diagnose each window to highlight the weaknesses of the current policy (Section \ref{sec:critic}).

\paragraph{Anchor selection and window construction}
\label{sec:anchor}
To avoid vulnerability across multiple frames, we construct a window of frames based uncertain decisions and poor grounding visual ranked by $\rho_t$. We first collect candidate \emph{anchors}
$\mathcal{T}_{\mathrm{anchor}}=\{t:\rho_t>\tau_s\}$ for a salience threshold $\tau_s$, and
merge anchors separated by at most $\Delta_{\max}$ frames into maximal intervals
$\mathcal{I}=\{[a_i,b_i]\}_{i=1}^{M}$, so that one emerging event is represented once. For each
interval we take the representative anchor $t_i^\star=\arg\max_{t\in[a_i,b_i]}\rho_t$ and build
a temporal window of left/right context lengths $\ell,r$ (indices clipped to $[1,T]$):
\begin{equation}
  W_i=\{z_t\}_{t=t_i^\star-\ell}^{\,t_i^\star+r}.
  \label{eq:window}
\end{equation}

\paragraph{Critic and failure descriptor}
\label{sec:critic}
We want to highlight not merely
whether the policy failed, but a structured account of \emph{how} the failure arises from
perception, uncertainty, control, and scene context. To this end we generate additional rationales capture the failure mode of the policy using windows. Specifically, VLM-based critic $C_\omega:\mathcal{W}\to\mathcal{Y}\times\mathcal{R}$ maps each window to a
behavior label and a structured rationale, $(y_i,r_i)=C_\omega(W_i)$, with label space
$\mathcal{Y}=\{\textsc{Reasonable},\textsc{Hesitant},\textsc{Unreasonable}\}$ denoting,
respectively, stable and contextually appropriate behavior, delayed/uncertain/oscillatory
behavior, and clearly unsafe or policy-inconsistent behavior. Importantly, The rationale $r_i$ records the
diagnosed mechanism (failure type, dominant visual evidence, temporal control pattern, relevant
actor interaction, and severity). We summarize the critic output into a compact \emph{failure
descriptor}
\begin{equation}
  \chi_i=(W_i,y_i,r_i),
  \label{eq:chi}
\end{equation}
which serves as the conditioning signal for generation.

\subsection{Failure-Conditioned Scenario Generation}
\label{sec:generation}

This stage instantiates the conditional generator $q_\phi(\xi\mid c)$ to generate critical scene for improving the policy using the failure descriptor as its conditioning context,
$c=\chi_i$. 

\paragraph{Constructive generator.}
Given $\chi_i$, an LLM planner proposes a structured scenario plan
$p_i\sim q_\phi(\,\cdot\mid\chi_i\,)$, $p_i\in\mathcal{P}$, specifying road configuration,
actor placement and motion, interaction timing, and environmental conditions. A deterministic
compiler $\Pi:\mathcal{P}\to\mathcal{C}$ translates the plan into an executable simulator
program $c_i=\Pi(p_i)$ (e.g., a Scenic, CARLA, or BeamNG script), which instantiates a
scenario $\xi_i=\mathrm{Exec}(c_i)\in\Xi$. The composition
$\chi_i\mapsto p_i\mapsto c_i\mapsto\xi_i$ defines a sampler from $q_\phi(\xi\mid\chi_i)$.

\paragraph{Feasibility and fidelity constraints}
Not every instantiated scenario is admissible. We define the \emph{executable-and-feasible}
subset through three predicates,
\begin{equation}
  \Xi_{\mathrm{exec}}=\left\{\xi\in\Xi:\ \textsc{Exe}(\xi)\wedge\textsc{Phys}(\xi)
                                          \wedge\textsc{Rule}(\xi)\right\},
  \label{eq:exec-set}
\end{equation}
where $\textsc{Exe}$, $\textsc{Phys}$, and $\textsc{Rule}$ denote simulator executability,
physical feasibility, and traffic-rule consistency, defined as follows.

\emph{Simulator executability} $\textsc{Exe}(\xi)$ holds when the scenario can be
successfully loaded and rolled out in BeamNG.tech without runtime faults. This requires that
all referenced map and level assets, vehicle models, and props exist in the simulator, that
every vehicle is spawned at a collision-free pose on a valid drivable surface, that the
requested agents and their attributes stay within the engine's supported limits, and that the
rollout proceeds for the full horizon without crashes, asset-loading failures, or vehicles
spawning into terrain or static geometry. Scenarios that fail to initialize, drop the
simulation step, or terminate prematurely due to engine errors are excluded.

\emph{Physical feasibility} $\textsc{Phys}(\xi)$ holds when the initial state and the
prescribed trajectories are consistent with the vehicle dynamics that BeamNG.tech resolves
through its soft-body, node--beam physics. Concretely, every trajectory must be realizable by
the simulated powertrain and chassis under bounded longitudinal acceleration and braking,
bounded steering and curvature, and tire--road friction limits, without commanding states that
the physics engine cannot produce (e.g., instantaneous teleportation, discontinuous heading,
overlapping vehicle volumes at $t{=}0$, or cornering speeds beyond the available grip).
Because BeamNG.tech enforces these dynamics natively, $\textsc{Phys}$ additionally screens out
rollouts in which a vehicle loses controllability in an unintended way (e.g., spurious rollover,
loss of traction, or structural deformation) that does not correspond to the diagnosed failure.

\emph{Traffic-rule consistency} $\textsc{Rule}(\xi)$ holds when the configuration of the
non-ego traffic and the environment is consistent with the prevailing traffic regulations, so
that any unsafe outcome can be attributed to the policy $\pi_\theta$ under test rather than to
an ill-posed world. We require that background vehicles respect lane assignments and travel
directions, obey signal and right-of-way conventions at intersections, and keep signage,
signals, and road topology mutually consistent on the chosen map. This excludes degenerate
scenarios in which a collision is unavoidable regardless of the ego policy (e.g., a vehicle
forced to drive the wrong way into oncoming traffic), ensuring that the diagnosed failure
reflects a genuine deficiency of $\pi_\theta$ rather than an adversarially broken environment.

\paragraph{Selection objective}
Conceptually, generation selects, among feasible scenarios, one whose induced policy behavior
best matches the target diagnosis:
\begin{multline}
  \hat{\xi}_i \;\in\; \arg\max_{\xi\in\Xi_{\mathrm{exec}}}\ \mathrm{Align}(\xi,\chi_i), \\
  \mathrm{Align}(\xi,\chi_i)=\mathbb{E}_{\tau\sim P_\theta^{\xi}}\!\left[\,m(\tau,\chi_i)\,\right],
  \label{eq:gen-objective}
\end{multline}
where $m(\tau,\chi_i)\in[0,1]$ scores how strongly a trajectory $\tau$ exhibits the failure
mechanism described by $\chi_i$ (for instance, the salience-weighted agreement between the
on-rollout critique and $\chi_i$). Equation~\eqref{eq:gen-objective} makes precise the sense in
which SPHINX is failure-targeted: it does not merely seek high loss $L(\theta;\xi)$, but a
\emph{specific}, diagnosed failure mode.

\paragraph{Practical realization}
Because $\Pi$ and the simulator are not differentiable and $\Xi_{\mathrm{exec}}$ is defined by
hard predicates, we approximate Equation \ref{eq:gen-objective} by constrained sampling with a
validate-and-repair loop. We draw candidates from $q_\phi(\cdot\mid\chi_i)$ and accept only
those that are feasible and fidelity-preserving:
\begin{equation}
  \hat{\xi}_i\sim q_\phi(\xi\mid\chi_i)
  \quad\text{s.t.}\quad
  \hat{\xi}_i\in\Xi_{\mathrm{exec}}\ \wedge\ \textsc{Fid}(\hat{\xi}_i,\chi_i)=1 .
  \label{eq:gen-sampling}
\end{equation}
A candidate that violates Equation \ref{eq:exec-set} is repaired under simulator constraints
(or rejected and resampled); a candidate that is feasible but loses the target failure
condition ($\textsc{Fid}=0$) is likewise rejected. This rejection/repair procedure is a
tractable surrogate for the constrained selection in Equation \ref{eq:gen-objective}, keeping
generation aligned with the diagnosed weakness rather than with diversity or loss alone.

\subsection{Closed-Loop Policy Retraining}
\label{sec:retraining}

The accepted scenarios are executed under the current policy to produce additional rollouts.
Let:
\begin{equation}
  \mathcal{D}_{\mathrm{gen}}=\left\{(\hat{\xi}_i,\tau_i,\chi_i)\right\}_{i=1}^{N},
  \qquad \tau_i\sim P_\theta^{\hat{\xi}_i}(\tau),
  \label{eq:gen-data}
\end{equation}
be the generated set. SPHINX then updates the policy with the base AV model's own training
protocol, augmenting the original data with the generated scenarios:
\begin{equation}
  \theta^{+}=\operatorname{Update}\!\left(\theta;\ \mathcal{D}_{\mathrm{orig}}
             \cup\mathcal{D}_{\mathrm{gen}}\right).
  \label{eq:update}
\end{equation}
The update may be behavior cloning, imitation learning, reinforcement learning, or
model-specific fine-tuning, depending on the base policy. This is precisely the policy step of
the adversarial objective Equation \ref{eq:update}, but with the inner generator restricted to the
feasible, failure-conditioned distribution of Equation \ref{eq:gen-sampling}: SPHINX reduces
$L(\theta;\xi)$ on scenarios that are executable \emph{and} aligned with previously diagnosed
weaknesses, rather than on arbitrary high-loss scenarios.

Evaluating $\theta^{+}$ yields new rollouts, new evidence $\mathcal{E}_{1:T}$, and new failure
descriptors $\{\chi_i\}$, which seed the next round of generation. This closes the loop: SPHINX
repeatedly \emph{explains} the current policy, \emph{explores} targeted scenarios derived from
that explanation, validates them in simulation, and retrains on the resulting data.

\begin{table*}[!tbp]
    \centering
    \caption{Scenario-wise post-training success rate across critical scenes and scenario-generation frameworks. All entries are reported as success rate (\%) evaluated over a broad and diverse set of challenging experimental scenarios. Overall, the base models remain the weakest, ChatScene provides modest gains, LLM-Attacker yields stronger improvements, and SPHINX achieves the most consistent performance across different critical scenes and driving backbones.}
    \label{tab:scenario_breakdown_scaled_down}
    \scriptsize
    \renewcommand{\arraystretch}{1.12}
    \setlength{\tabcolsep}{3.2pt}
    \resizebox{\textwidth}{!}{%
    \begin{tabular}{lcccccccccccccccc}
        \toprule
        \multirow{2}{*}{\textbf{Critical Scene}} 
        & \multicolumn{4}{c}{\textbf{Base Performance}} 
        & \multicolumn{4}{c}{\textbf{LLM-Attacker}} 
        & \multicolumn{4}{c}{\textbf{ChatScene}} 
        & \multicolumn{4}{c}{\textbf{SPHINX (Ours)}} \\
        \cmidrule(lr){2-5}\cmidrule(lr){6-9}\cmidrule(lr){10-13}\cmidrule(lr){14-17}
        & \textbf{DAVE-2} & \textbf{TransFuser} & \textbf{TCP} & \textbf{RAP-ResNet}
        & \textbf{DAVE-2} & \textbf{TransFuser} & \textbf{TCP} & \textbf{RAP-ResNet}
        & \textbf{DAVE-2} & \textbf{TransFuser} & \textbf{TCP} & \textbf{RAP-ResNet}
        & \textbf{DAVE-2} & \textbf{TransFuser} & \textbf{TCP} & \textbf{RAP-ResNet} \\
        \midrule
        Cut-in Left
        & 22.74 & 39.82 & 29.75 & 38.84
        & 55.30 & 61.31 & 69.67 & 62.57
        & 38.13 & 47.37 & 55.00 & 45.63
        & \textbf{80.42} & \textbf{86.93} & \textbf{93.18} & \textbf{87.26} \\

        Sudden Brake
        & 31.85 & 29.78 & 46.74 & 31.74
        & 69.24 & 78.54 & 79.54 & 69.63
        & 55.46 & 61.54 & 61.72 & 54.75
        & \textbf{93.62} & \textbf{87.14} & \textbf{99.73} & \textbf{93.41} \\

        Parked Vehicle Pullout
        & 15.55 & 31.65 & 38.18 & 30.58
        & 46.49 & 62.90 & 60.86 & 71.13
        & 31.22 & 45.95 & 46.50 & 55.96
        & \textbf{86.81} & \textbf{93.52} & \textbf{87.06} & \textbf{93.87} \\

        Intersection Collision
        & 16.62 & 23.29 & 31.22 & 22.84
        & 53.95 & 62.34 & 70.50 & 61.05
        & 39.13 & 47.22 & 53.53 & 47.01
        & \textbf{87.12} & \textbf{86.64} & \textbf{93.74} & \textbf{99.28} \\

        Wrong-way Driver
        & 23.64 & 30.42 & 39.79 & 30.12
        & 62.58 & 61.30 & 69.98 & 78.41
        & 45.70 & 47.01 & 55.38 & 61.33
        & \textbf{80.73} & \textbf{87.35} & \textbf{93.49} & \textbf{86.82} \\
        \midrule
        \textbf{Average}
        & 22.08 & 30.99 & 37.14 & 30.82
        & 57.51 & 65.28 & 70.11 & 68.16
        & 41.93 & 49.82 & 54.43 & 52.94
        & \textbf{85.74} & \textbf{88.32} & \textbf{93.44} & \textbf{92.13} \\
        \bottomrule
    \end{tabular}%
    }
\end{table*}

\section{Experiments}

\subsection{Baselines and Settings} 

\textbf{Autonomous Vehicle Models.} We evaluate our framework on four representative autonomous vehicle (AV) models spanning diverse architectural paradigms. DAVE-2  \cite{bojarski2016endendlearningselfdriving} (NVIDIA 2016) is a classical end-to-end CNN that maps raw images to control commands via behavior cloning, serving as a baseline with high sensitivity to perceptual noise and distribution shift. MILE \cite{hu2022model} (NeurIPS 2022) introduces a model-based imitation learning approach that leverages bird’s-eye-view (BEV) representations and latent dynamics to provide more structured and semantically meaningful reasoning. TransFuser \cite{chitta2022transfuser} (IEEE Trans. Pattern Anal. Mach. Intell., 2022) further advances this line by employing a transformer-based architecture to fuse multi-modal camera and LiDAR inputs, enabling strong performance in complex traffic through long-range and cross-modal interactions. Finally, RAP-ResNet \cite{feng2026rap} (ICLR 2026) represents a modern robust planning framework that integrates 3D rasterization and counterfactual reasoning, achieving state-of-the-art performance and improved sim-to-real generalization on benchmarks such as Bench2Drive \cite{jia2024bench2drive}. This diverse selection allows us to systematically evaluate whether our framework can operate in a model-agnostic and plug-and-play manner without relying on any specific architectural assumption.

\textbf{Scene Generation Settings and Metrics.} Our framework is designed in a plug-and-play manner with respect to the explainer module, allowing it to adapt to different architectures. In this work, we instantiate the explainer using CRAFT \cite{fel2023craft3} due to its practical implementability and its ability to provide concept-level explanations that capture both what the model attends to and where it focuses. We further compare our framework against two state-of-the-art scenario generation methods, ChatScene~\cite{zhang2024chatscene} (CVPR 2024) and LLM-Attacker ~\cite{mei2025llm,peng2025ldscenellmguideddiffusioncontrollable} (IEEE TITS 2025), both of which leverage large language models to synthesize driving scenarios through procedural reasoning. To ensure a fair comparison, all methods are evaluated under a unified set of test scenarios constructed by aggregating the critical scenarios generated by each method. Performance is assessed using a standard metric, which is the Success Rate (SR), defined as the fraction of scenarios completed without infractions; The metric is computed over the same set of evaluation scenarios to ensure consistency across methods

\begin{figure*}[!tbp]
    \centering
    \includegraphics[width=\linewidth]{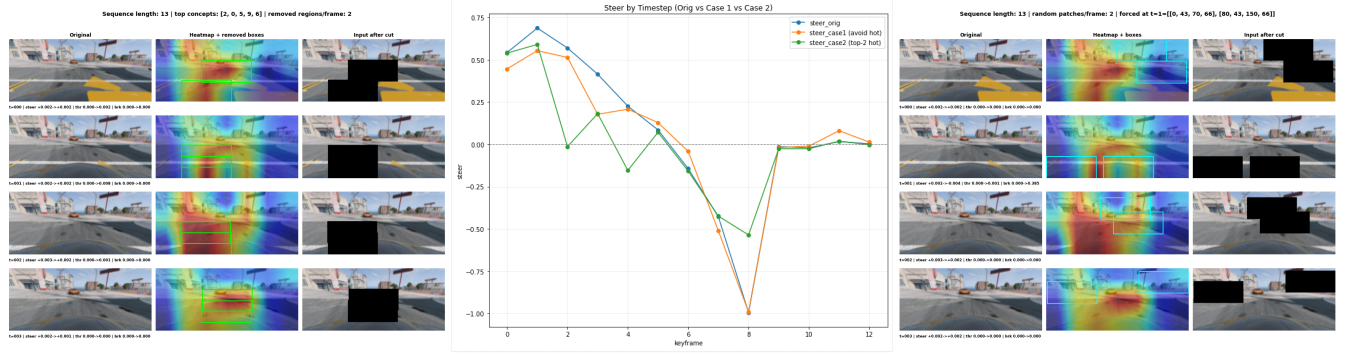}
    \caption{Spatial faithfulness of the extracted concepts through a masking-based causal test on a $13$-frame steering sequence (top concepts $[2,0,5,9,6]$). \textbf{Left:} for each frame we localize the top-concept (heat-dominant) regions from the aggregated concept heatmap (green boxes) and mask them out of the input (\emph{Input after cut}). \textbf{Middle:} the resulting per-timestep steering trajectories, comparing the original prediction (\texttt{steer\_orig}) with two interventions---avoiding the hot regions (\texttt{steer\_case1}) and removing the top-$2$ hot regions (\texttt{steer\_case2}). Removing the concept-relevant regions produces the largest deviation from the original trajectory, particularly over the early-to-mid keyframes, while all three curves converge during the strong committed maneuver near keyframe~$8$ (\texttt{steer}~$\approx-1.0$). \textbf{Right:} a control condition in which the same number of \emph{randomly located} patches (cyan boxes, $2$ per frame) are masked instead of the concept regions; this perturbation leaves the steering trajectory essentially unchanged. The contrast between the concept-based and random ablations indicates that the regions highlighted by the extracted concepts are causally responsible for the model's steering, rather than being incidental correlates.}
    \vspace{-2mm}
    \label{fig:spatial_ablation_examples}
\end{figure*}

\subsection{Comparative Analysis of Scenario-Generation Methods}
Based on Table~\ref{tab:scenario_breakdown_scaled_down}, SPHINX consistently demonstrates significant performance improvements over both LLM-Attacker and ChatScene across all evaluated AV models, ranging from classical end-to-end CNN architectures such as DAVE-2 to more advanced hybrid and transformer-based systems including TransFuser, TCP, and RAP-ResNet. Importantly, this improvement is not limited to aggregate metrics but is consistently observed across individual scenario types. For instance, across all evaluated scenarios, SPHINX consistently achieves high success rates (typically above 80\% and reaching up to 100\%) on every model, whereas baseline methods remain below 70\% in most cases. This gap persists across both simpler scenarios such as \textit{Cut-in Left} and more complex cases including \textit{Intersection Collision} and \textit{Wrong-way Driver}, highlighting a consistent and substantial performance difference. Since all methods are evaluated on the same set of scenarios, the observed performance gap can be directly attributed to the quality of scenario generation rather than differences in test exposure or data distribution.

These results lead to two key insights. First, SPHINX outperforms all baseline methods (including the base models, ChatScene, and LLM-Attacker) because these results can be attributed to a fundamental difference in how scenarios are generated. Prior methods such as ChatScene and LLM-Attacker rely primarily on generic priors from language or vision-language models, producing scenarios that may induce failures but are not necessarily aligned with the specific weaknesses of the evaluated policy. In contrast, SPHINX first explains the policy’s behavior by extracting model-internal failure evidence, and then conditions scenario generation on these explanations. By grounding generation in interpretable failure signals, SPHINX produces scenarios that directly target the underlying failure mechanisms, leading to more informative training data and more effective policy improvement.

Second, the consistently high performance of SPHINX across all scenarios and models suggests strong generalization. The retrained policies are able to handle a diverse range of scenario types, from relatively simple interactions such as cut-in events to more complex situations such as intersection conflicts and wrong-way driving. This behavior indicates that SPHINX does not overfit to a narrow subset of scenarios, but instead improves robustness across a broader failure distribution. 

\begin{figure}[!tbp]
    \centering
    \includegraphics[width=\linewidth]{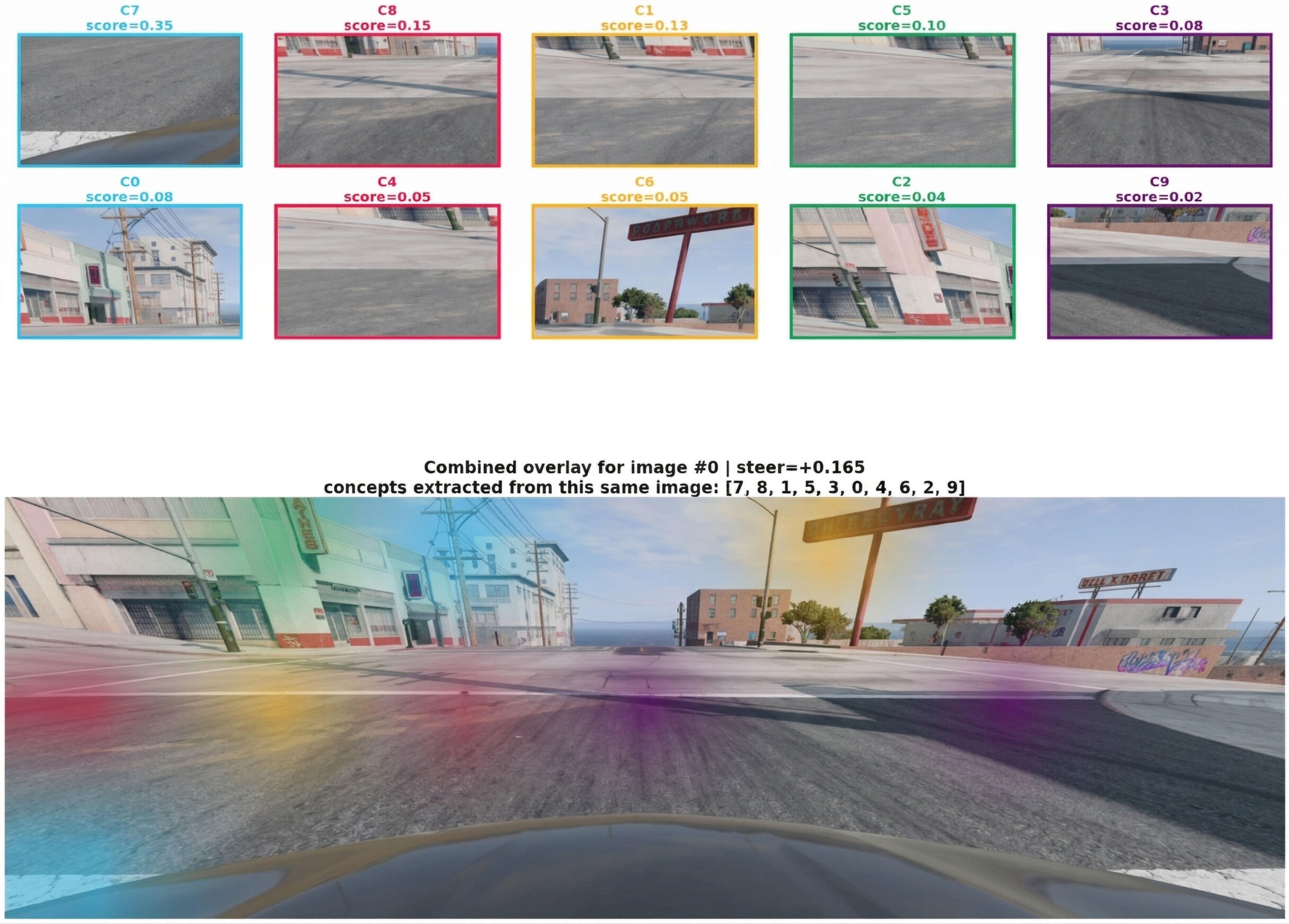}
    \caption{Visualization of the extracted foundational concepts ($C_0$--$C_9$) and their spatial projection for the steering-focused subset. Top: the top ten visual concepts identified by CRAFT, each assigned a distinct color code and summarized via representative crops. Bottom: the corresponding semantic heatmap overlaid on the original driving frame, illustrating the model's visual grounding during a specific steering output (\textit{steer = +0.165}).}
    \label{fig:concept_extraction}
\end{figure}

\subsection{Explanation Faithfulness}

To evaluate the faithfulness of the extracted explanation signals, we conduct both qualitative analysis and causal validation through spatial ablation experiments. We first examine the concept-level heatmaps shown in 
Fig.~\ref{fig:concept_extraction}. The high-intensity regions identified by the XAI module consistently align with semantically relevant areas for decision-making, such as lane direction and intersection structure. Correspondingly, the model produces a steering command of $\text{steer} = +0.165$, indicating a left-turn behavior that is consistent with the highlighted regions. This observation suggests a strong alignment between the explanation signals and the policy’s control decisions.

To further assess whether these regions are causally relevant, we perform controlled masking experiments on a $13$-frame steering sequence, as illustrated in Fig.~\ref{fig:spatial_ablation_examples}. For each frame, we localize the top-concept (heat-dominant) regions from the aggregated concept heatmap and mask them out of the input, then re-run the policy and record the resulting steering trajectory. We consider two interventions: avoiding the hot regions (\texttt{case 1}) and removing the top-$2$ hot regions (\texttt{case 2}). As shown in the middle panel of Fig.~\ref{fig:spatial_ablation_examples}, removing the concept-relevant regions produces a clear deviation from the original steering trajectory, most pronounced over the early-to-mid keyframes where the decision is not yet dominated by a single committed maneuver; near keyframe~$8$, all curves converge as the policy executes a strong turn (\texttt{steer}~$\approx-1.0$) that is driven by the overall scene geometry. To verify that this effect is specific to the concept regions rather than a generic consequence of occlusion, we introduce a control condition (right panel) in which the same number of \emph{randomly located} patches is masked in each frame. This random ablation leaves the steering trajectory essentially unchanged. The contrast between the two conditions demonstrates that the regions identified by the extracted concepts contain information that is causally necessary for the model's decision, whereas equally sized random regions are not. Taken together, these results provide strong empirical evidence that our extracted explanations are not merely descriptive but faithfully capture the causal factors underlying the model's behavior. This level of faithfulness enables the explanation signals to serve as a reliable conditioning mechanism for downstream components in the SPHINX pipeline.

\subsection{From Failure Mechanism to Corrected Behavior}

\begin{figure*}[!tbp]
    \centering 
    \begin{subfigure}[t]{0.48\textwidth} 
        \centering 
        \includegraphics[width=\linewidth]{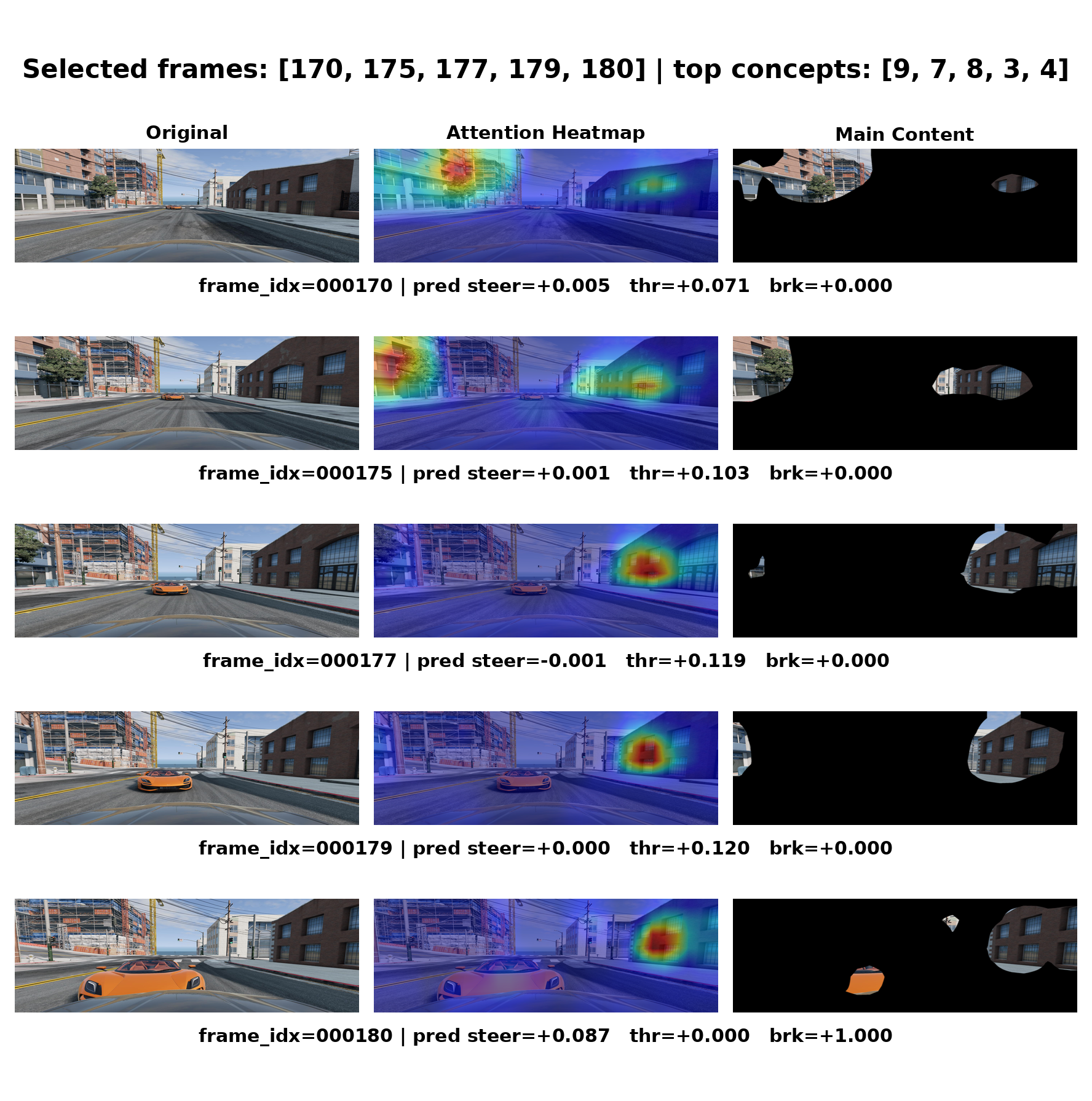} 
        \caption{Late detection and hesitant evasive response to an oncoming vehicle.} 
        \label{fig:vlm_failure_oncoming} 
    \end{subfigure} 
    \hfill 
    \begin{subfigure}[t]{0.48\textwidth} 
        \centering 
        \includegraphics[width=\linewidth]{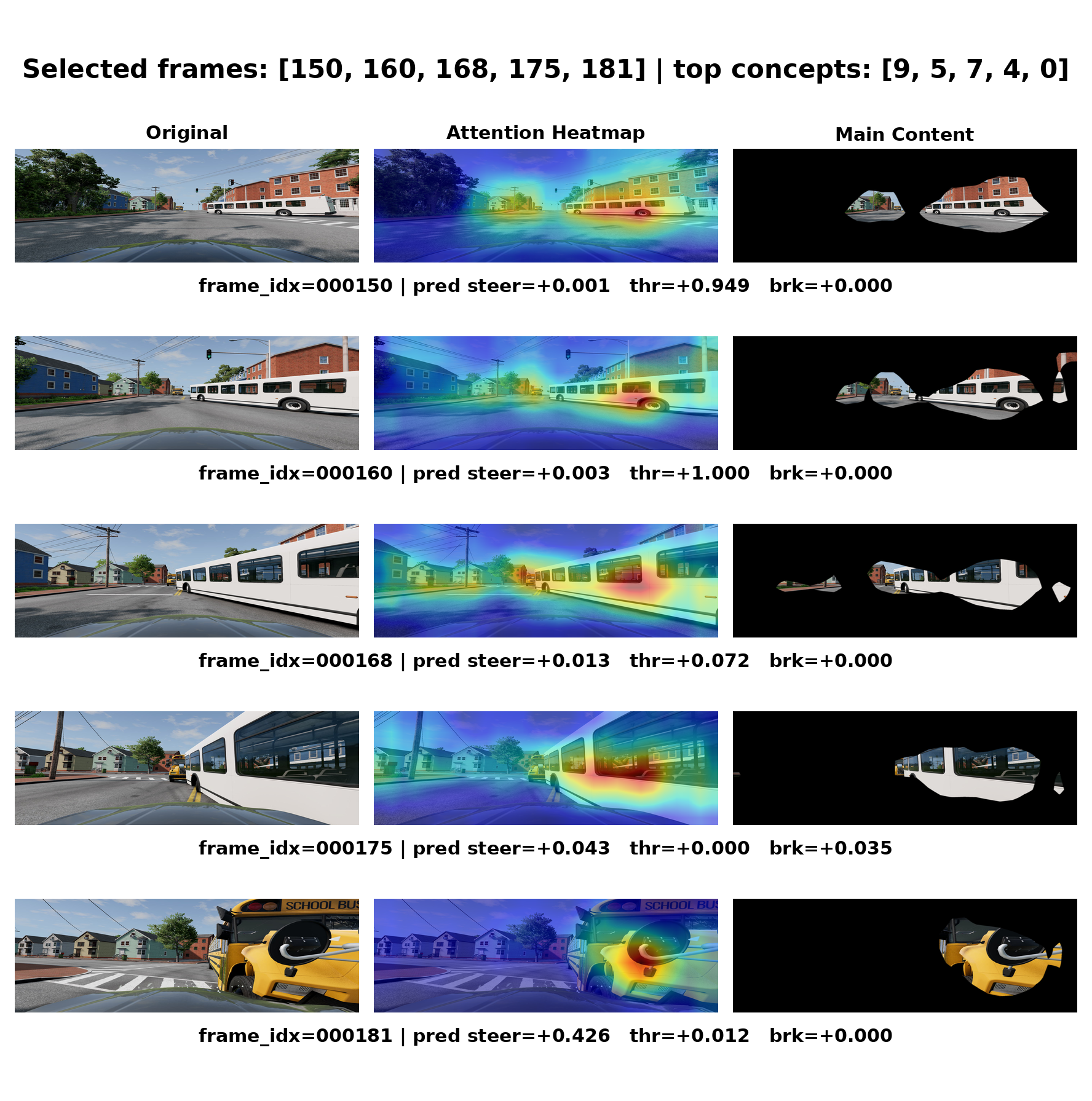} 
        \caption{Hesitant and late steering/braking when a vehicle turns across the ego path from the right side of the intersection.} 
        \label{fig:vlm_failure_bus} 
    \end{subfigure} 
    \caption{Representative temporal failure modes analyzed by the VLM critic. \textbf{Left:} the ego policy initially attends to lane and roadside cues, detects the oncoming vehicle only at close range, and reacts too late to avoid collision. \textbf{Right:} in the bus-collision case, a vehicle turns into the ego route from the right side of the intersection, but the policy responds with hesitant and late steering/braking instead of slowing down and avoiding proactively.}   \label{fig:vlm_failure_taxonomy} 
    \vspace{-2mm}
\end{figure*}

\begin{figure*}[!tbp]
    \centering
    \begin{subfigure}[t]{0.48\textwidth}
        \centering
        \includegraphics[width=\linewidth]{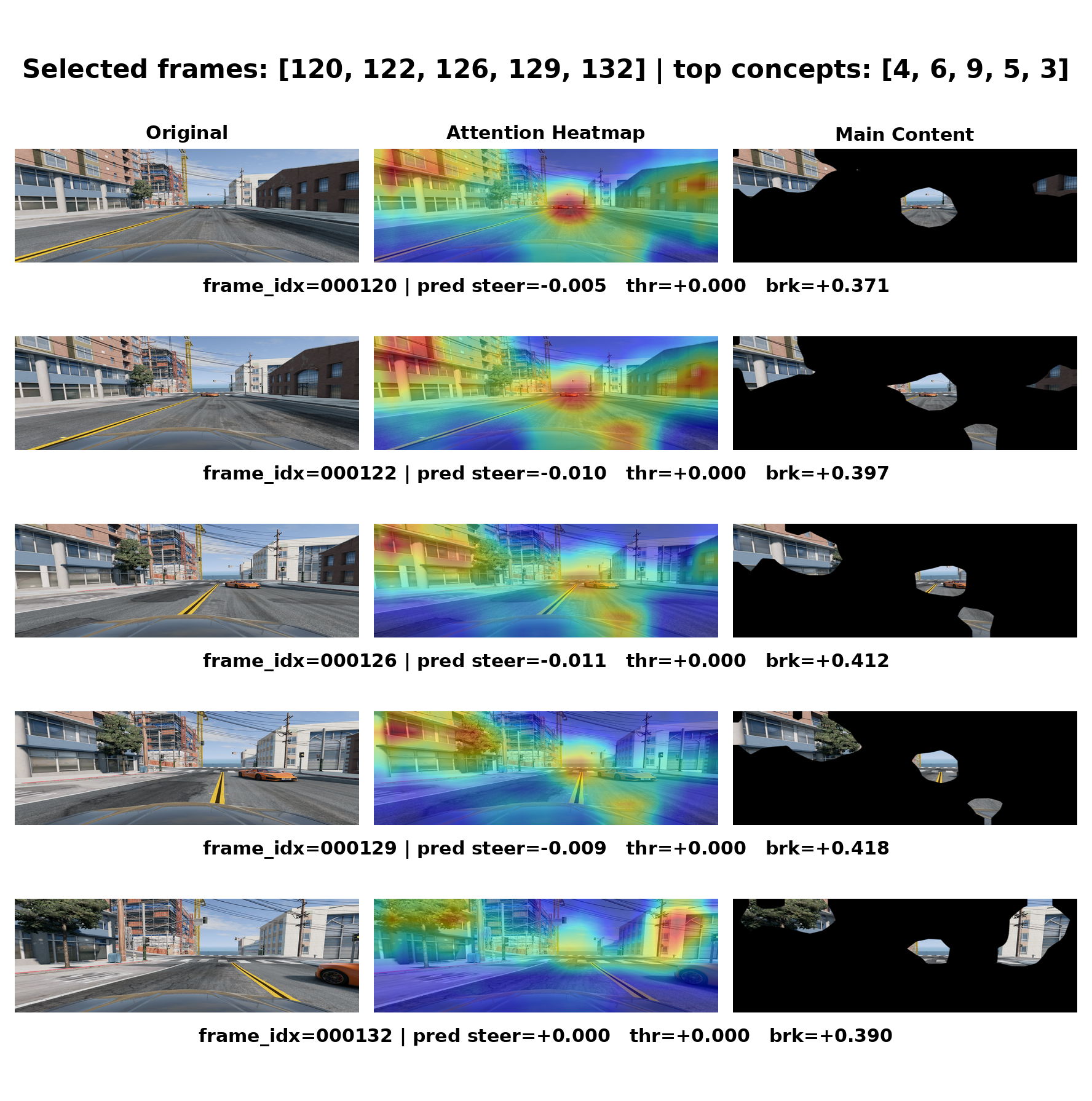}
        \caption{After fine-tuning with SPHINX, the model focuses on the approaching wrong-way vehicle and produces a smoother avoidance response across frames 120--132: the throttle is fully released, braking is applied early and consistently, and steering stays small and controlled.}
        \label{fig:after_training_sphinx}
    \end{subfigure}
    \hfill
    \begin{subfigure}[t]{0.48\textwidth}
        \centering
        \includegraphics[width=\linewidth]{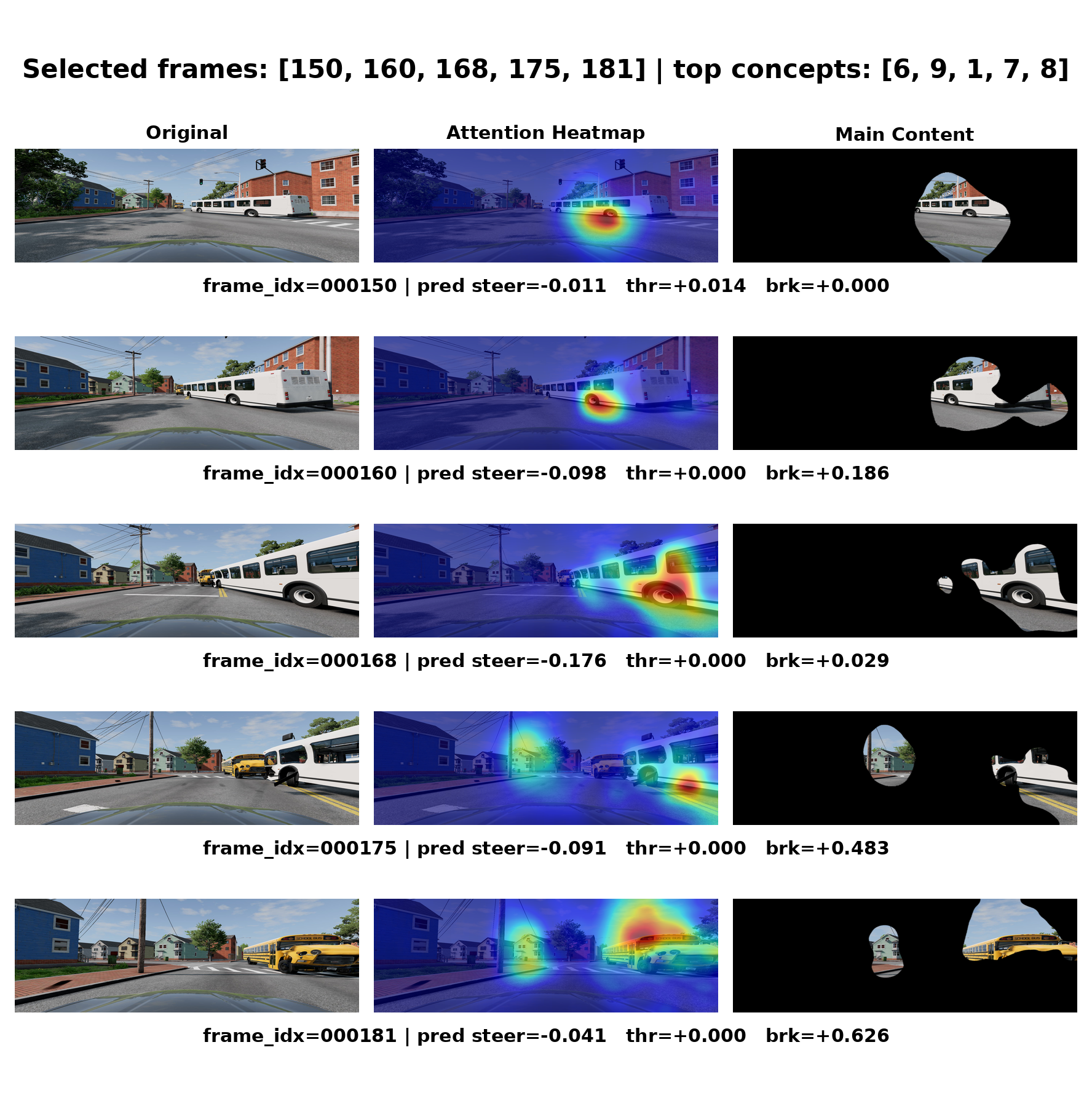}
        \caption{After fine-tuning with SPHINX, the model proactively slows down when the right-side crossing vehicle becomes relevant and steers to avoid the nearby actor.}
        \label{fig:after_training_bus_sphinx}
    \end{subfigure}
    \caption{Qualitative examples after fine-tuning with the SPHINX. \textbf{Left:} in the wrong-way case, the model focuses on the approaching vehicle and produces a controlled response over frames 120--132, releasing the throttle and braking early and consistently while keeping steering small. \textbf{Right:} in the bus-collision scenario, the retrained model releases the throttle, applies progressively stronger braking, and steers away from the vehicle turning across the ego path from the right side of the intersection.}
    \label{fig:after_training_examples_sphinx}
    \vspace{-2mm}
\end{figure*}

After establishing the faithfulness of our explanations, we identify a key insight: the observed failures are not caused by missing visual information alone, but by how the model converts that information into temporally coherent control. Fig.~\ref{fig:vlm_failure_oncoming} illustrates this behavior in the wrong-way interaction. Before retraining, the policy observes the approaching vehicle but does not form a stable avoidance plan. Across the selected frames \texttt{170}, \texttt{175}, \texttt{177}, \texttt{179}, and \texttt{180}, attention remains mixed between the road, surrounding structures, and the actor, while the steering stays essentially neutral (\texttt{+0.005}, \texttt{+0.001}, \texttt{-0.001}, \texttt{+0.000}, \texttt{+0.087}) and the throttle is held open (\texttt{+0.071}, \texttt{+0.103}, \texttt{+0.119}, \texttt{+0.120}, \texttt{+0.000}) with no braking until the very last frame. Only when the collision is imminent does the model react, with a full emergency brake (\texttt{brk=+1.000} at frame \texttt{180})---a reaction that occurs after the available response margin has already disappeared. After retraining with SPHINX (Fig.~\ref{fig:after_training_sphinx}), the behavior becomes grounded and consistent with the displayed telemetry. Over frames \texttt{120}, \texttt{122}, \texttt{126}, \texttt{129}, and \texttt{132}, the model fully releases the throttle (\texttt{thr=+0.000} throughout) and applies early, sustained braking (\texttt{+0.371}, \texttt{+0.397}, \texttt{+0.412}, \texttt{+0.418}, \texttt{+0.390}) while keeping steering small and controlled (\texttt{-0.005}, \texttt{-0.010}, \texttt{-0.011}, \texttt{-0.009}, \texttt{+0.000}) rather than deferring to an abrupt last-moment maneuver. The heatmaps and isolated main-content panels show that attention shifts toward the approaching orange vehicle and the local conflict region rather than remaining dominated by background structure. This demonstrates that SPHINX does not merely expose where the model looks; it turns that evidence into targeted training data that improves the timing and consistency of the downstream action.

The second case, shown in Fig.~\ref{fig:vlm_failure_bus}, corresponds to a bus-collision scenario in which the original policy exhibits \emph{hesitant and late steering and braking}. The critical hazard is not simply a static object: a vehicle emerges from the right side of the intersection and turns across the ego route while the ego vehicle is still moving forward. Before retraining, across frames \texttt{150}, \texttt{160}, \texttt{168}, \texttt{175}, and \texttt{181}, the model keeps the throttle near or at maximum as the crossing vehicle approaches (\texttt{thr=+0.949}, \texttt{+1.000}) and barely brakes (\texttt{brk=+0.000} for most of the window, peaking at only \texttt{+0.035}), committing to no avoidance direction until the conflict is imminent. It then produces a sharp, last-moment steer (\texttt{steer=+0.426} at frame \texttt{181}), leaving too little time to resolve the conflict. After SPHINX retraining (Fig.~\ref{fig:after_training_bus_sphinx}), the failure mode is corrected. Over the same frames, the model releases the throttle early (\texttt{thr=+0.000} after the first frame) and applies progressively stronger braking as the crossing vehicle becomes relevant (\texttt{brk=+0.186}, \texttt{+0.029}, \texttt{+0.483}, \texttt{+0.626}), while steering away from the actor with a smooth, increasing left command (\texttt{steer=-0.098}, \texttt{-0.176}, \texttt{-0.091}, \texttt{-0.041}). The attention heatmaps and isolated main-content panels become more aligned with the approaching vehicle and its conflict region. This indicates an improved perception-to-control mapping: the retrained policy not only detects the hazard, but also decelerates and steers away from it before collision risk becomes unrecoverable.

\begin{table*}[!tbp]
    \centering
    \caption{Case study for the wrong-way collision sequence in Fig.~\ref{fig:vlm_failure_oncoming} and the corrected behavior in Fig.~\ref{fig:after_training_sphinx}. The table illustrates how SPHINX decomposes a single high-level failure---the ego vehicle approaching the intersection too fast and colliding with a wrong-way vehicle---into a set of concrete, independently fixable sub-problems, and, for each one, what kind of corrective training data is synthesized to address it.}
    \label{tab:case_study_wrongway_collision}
    \small
    \renewcommand{\arraystretch}{1.4}
    \setlength{\tabcolsep}{5pt}
    \begin{tabularx}{\linewidth}{
        >{\RaggedRight\arraybackslash}p{0.16\linewidth}
        >{\RaggedRight\arraybackslash}X
        >{\RaggedRight\arraybackslash}X
    }
        \toprule
        \textbf{Sub-problem} & \textbf{Diagnosed deficiency in the original policy} & \textbf{Corrective data synthesized by SPHINX} \\
        \midrule

        \textbf{(S1) No speed regulation on approach}
        & The ego vehicle enters the intersection region at high speed and does not slow down, leaving insufficient margin to react once a conflict appears.
        & Scenarios that preserve the same intersection geometry but require early deceleration, with rollouts labeled to reward throttle release and anticipatory braking on approach \texttt{(thr$\approx$0, brk$>$0 before the conflict region)}. \\

        \textbf{(S2) No attention to the incoming actor}
        & Visual attention is spread across road surface, lane markings, and roadside buildings; the approaching vehicle does not dominate the evidence used for control.
        & Counterfactual scenes that keep the wrong-way actor but vary static background cues, paired with attention/saliency targets that concentrate on the moving vehicle and the local conflict region. \\

        \textbf{(S3) Hesitant, unstable decision}
        & The policy does not commit early to a single safe direction; it oscillates and only reacts strongly once the actor is already close, producing a late evasive maneuver.
        & Trajectories that demonstrate early, consistent directional commitment, with smooth steering profiles rewarded over late, abrupt corrections. \\

        \textbf{(S4) Threat not recognized as an avoidance trigger}
        & Even when the wrong-way vehicle is visible, the policy does not treat it as a safety-critical hazard requiring an evasive response.
        & Fidelity-preserving variants in which the wrong-way vehicle is the dominant threat, with critique labels ($y\neq\textsc{Reasonable}$) tying its presence to a required, timely avoidance action. \\

        \midrule
        \textbf{Combined outcome}
        & \multicolumn{2}{>{\RaggedRight\arraybackslash}X}{%
        Jointly addressing \textbf{(S1)}--\textbf{(S4)} yields the corrected behavior over frames \texttt{120}--\texttt{132}: the retrained policy grounds its response on the approaching orange vehicle, fully releases the throttle \texttt{(thr=+0.000)}, and applies early, sustained braking \texttt{(brk$\approx$+0.37 to +0.42)} with small, controlled steering before the available margin disappears (Fig.~\ref{fig:after_training_sphinx}).} \\

        \bottomrule
    \end{tabularx}
\end{table*}

\subsection{Case Study: From Failure Evidence to Targeted Scenario Correction}
\label{sec:case_study_wrongway_collision}

Table~\ref{tab:case_study_wrongway_collision}, Fig.~\ref{fig:vlm_failure_oncoming}, and Fig.~\ref{fig:after_training_sphinx} illustrates how SPHINX converts a temporal wrong-way collision failure into a targeted scenario correction. The figures provide the raw evidence: synchronized RGB frames, attention heatmaps, isolated main-content regions, and predicted control signals. The table then converts this evidence into a mechanism-level diagnosis. In this case, the failure is not that the wrong-way vehicle is completely invisible; rather, the policy is \emph{hesitant} and translates the visual cue into a safe avoidance action too late.

\textbf{Failure Type.}
The wrong-way case is a dynamic head-on conflict. The approaching vehicle becomes visible before the collision point, but the original policy does not produce an early and stable avoidance plan; it holds the throttle open with near-neutral steering and brakes only with a full emergency stop at the last frame. The failure is therefore characterized as \textbf{hesitant and late evasive response}: the model delays action until the actor is already close. SPHINX transforms this descriptive failure into a training objective: prioritize the wrong-way actor earlier, release the throttle, apply braking early as the gap closes, and steer away while sufficient maneuvering space remains.

\textbf{Event Description.}
As shown in Fig.~\ref{fig:after_training_sphinx}, the corrected sequence spans frames \texttt{120}, \texttt{122}, \texttt{126}, \texttt{129}, and \texttt{132}. The orange wrong-way vehicle moves from a distant forward hazard into a close conflict region. The retrained model tracks this progression: attention and main-content masks increasingly include the actor and its surrounding conflict area, while the predicted actions remain controlled rather than abrupt.

\textbf{VLM / Critic Finding.}
The VLM critic identifies the original policy state as \textbf{HESITANT}: the actor is visible, but the model does not commit early enough to a safe avoidance strategy. After SPHINX retraining, Fig.~\ref{fig:after_training_sphinx} shows a more reasonable response: the model focuses on the approaching vehicle, releases the throttle, and applies braking early rather than waiting until the interaction becomes close.

\textbf{Attention Pattern Based On Heat Map.}
The post-correction heatmaps in Fig.~\ref{fig:after_training_sphinx} show attention concentrated near the approaching wrong-way vehicle and the local conflict region. The \textit{Main Content} panels preserve the actor and nearby drivable context while suppressing large portions of irrelevant background. This supports the interpretation that the retrained policy grounds its control in the safety-critical actor rather than in static scene texture alone.

\textbf{Telemetry Evidence.}
The displayed telemetry matches the corrected behavior. The throttle is fully released across the selected frames \texttt{(thr=+0.000 throughout)} and braking is applied early and consistently \texttt{(+0.371, +0.397, +0.412, +0.418, +0.390)} rather than as an abrupt last-moment reaction. Steering stays small and controlled \texttt{(-0.005, -0.010, -0.011, -0.009, +0.000)}, in contrast to the original policy, which held the throttle open and triggered a full emergency brake \texttt{(brk=+1.000)} only at the final frame. These signals indicate that SPHINX improves both response timing and action consistency.

\textbf{Environment and Actor Dynamics.}
The scene contains competing static cues, including road texture, lane direction, and buildings on both sides. These cues can distract the model from the approaching actor if the training data does not emphasize the causal role of the hazard. SPHINX addresses this by curating scenario evidence around the moving vehicle and its conflict region, making the actor dynamics central to the learned response.

Overall, the case study shows how SPHINX moves from failure evidence to intervention. The heatmaps and main-content panels identify what visual evidence should drive the policy; the telemetry verifies whether that evidence becomes timely control; and the curated scenario converts the diagnosis into training data. The resulting policy is not merely better at recognizing the wrong-way vehicle in a static frame. It becomes better at using the approaching actor as a temporal cue for low-throttle, controlled-steering, and timely-braking behavior before collision risk becomes unrecoverable.
\FloatBarrier
\section{Conclusion}
In this paper, we presented SPHINX, a novel closed-loop framework for adversarial driving scenario synthesis that operationalizes the 'first explain, then explore' principle. By transitioning Explainable AI (XAI) from a passive diagnostic tool to an active generative component, SPHINX effectively bridges the gap between internal failure mechanisms and scenario generation. Crucially, the framework is architecture-agnostic, demonstrating its efficacy across a wide range of AV models, ranging from traditional CNN-based architectures to modern Transformer-based systems. Furthermore, SPHINX features a modular design that allows for 'plug-and-play' integration with various XAI explainers, ensuring flexibility and adaptability to evolving interpretability techniques. Experimental evaluations demonstrate that our approach not only outperforms baseline methods in failure detection but also exhibits significant generalization and diagnostic precision. The ability to condition scenario generation on concept evidence and uncertainty provides a structured pathway for targeted retraining, ultimately enhancing the robustness of autonomous systems.


\FloatBarrier




%





\ifCLASSOPTIONcaptionsoff
  \newpage
\fi





\bibliographystyle{IEEEtran}
\bibliography{IEEEabrv,Bibliography,references}

\makeatletter
\newlength{\bioW}\newlength{\bioH}
\newlength{\bio@dx}\newlength{\bio@dy}\newlength{\bio@tw}\newlength{\bio@th}
\newsavebox{\bio@raw}\newsavebox{\bio@box}\newsavebox{\bio@cl}
\setlength{\bioW}{1in}\setlength{\bioH}{1.25in} 
\newcommand{\bioPhoto}[1]{%
  \sbox{\bio@raw}{\includegraphics{#1}}
  \setlength{\bio@tw}{\wd\bio@raw}\setlength{\bio@th}{\ht\bio@raw}%
  \pgfmathsetmacro{\bio@sfw}{\bioW/\bio@tw}%
  \pgfmathsetmacro{\bio@sfh}{\bioH/\bio@th}%
  \pgfmathsetmacro{\bio@sf}{max(\bio@sfw,\bio@sfh)}%
  \sbox{\bio@box}{\scalebox{\bio@sf}{\usebox{\bio@raw}}}%
  \setlength{\bio@dx}{\dimexpr(\wd\bio@box-\bioW)/2\relax}%
  \setlength{\bio@dy}{\dimexpr(\ht\bio@box-\bioH)/2\relax}%
  \sbox{\bio@cl}{\clipbox{\bio@dx \bio@dy \bio@dx \bio@dy}{\usebox{\bio@box}}}%
  \makebox[\bioW][c]{\raisebox{0pt}[\bioH][0pt]{\usebox{\bio@cl}}}%
}
\makeatother
\vspace{-10mm}
\begin{IEEEbiography}[{\bioPhoto{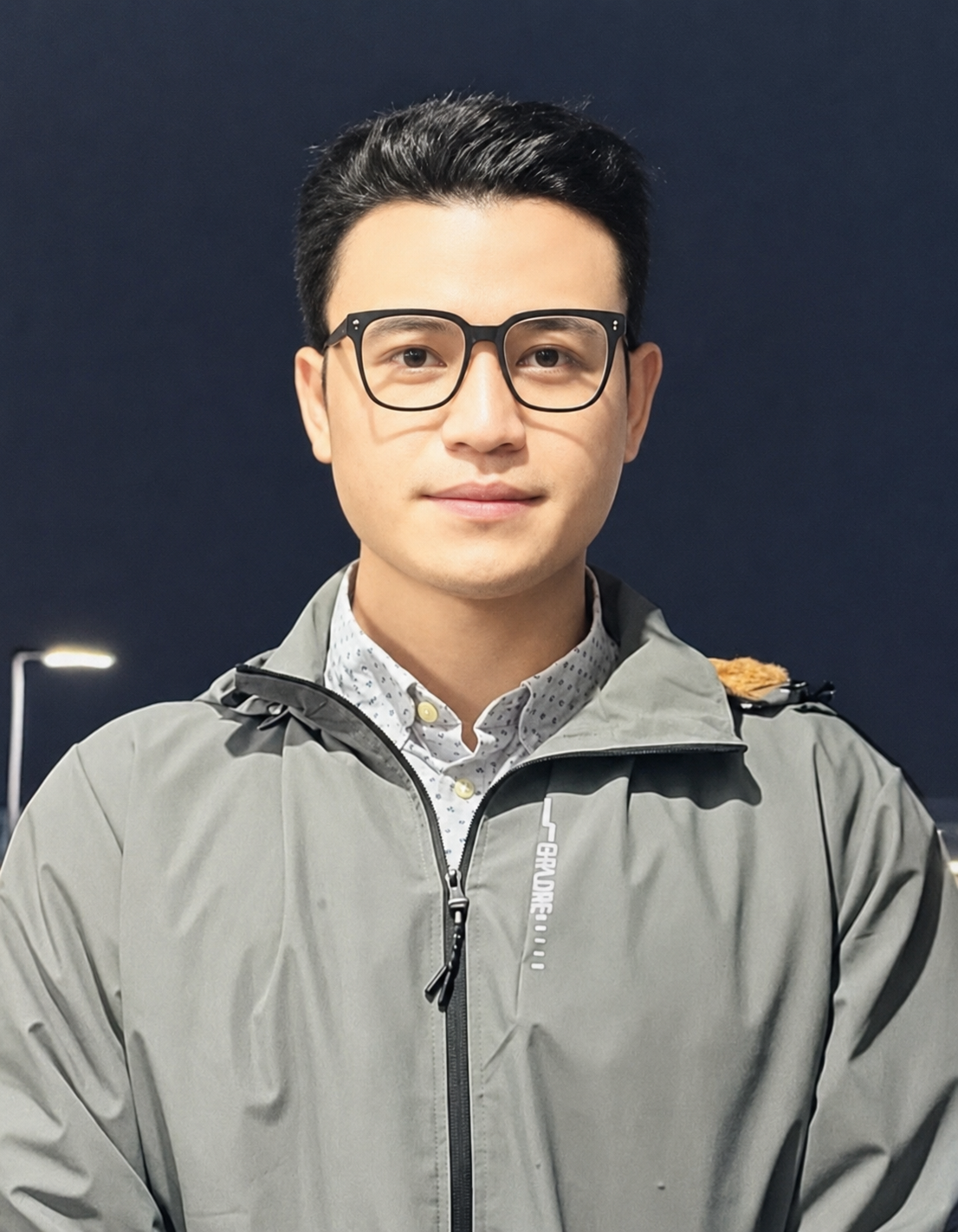}}]{Nguyen Do}
(Nguyen Hoang Khoi Do) received the B.Eng. degree in telecommunications from the Posts and Telecommunications Institute of Technology (PTIT), Hanoi, Vietnam, in 2021, and the M.Sc. degree in information systems from the same institution in 2024. He is currently working toward the Ph.D. degree with the Department of Computer and Information Science and Engineering, University of Florida, Gainesville, FL, USA, under the supervision of Prof. My T. Thai. His research interests include reinforcement learning, exploration algorithms in dynamic systems, generative AI, and mixture-of-experts models.
\end{IEEEbiography}
\vspace{-10mm}
 
\begin{IEEEbiography}[{\bioPhoto{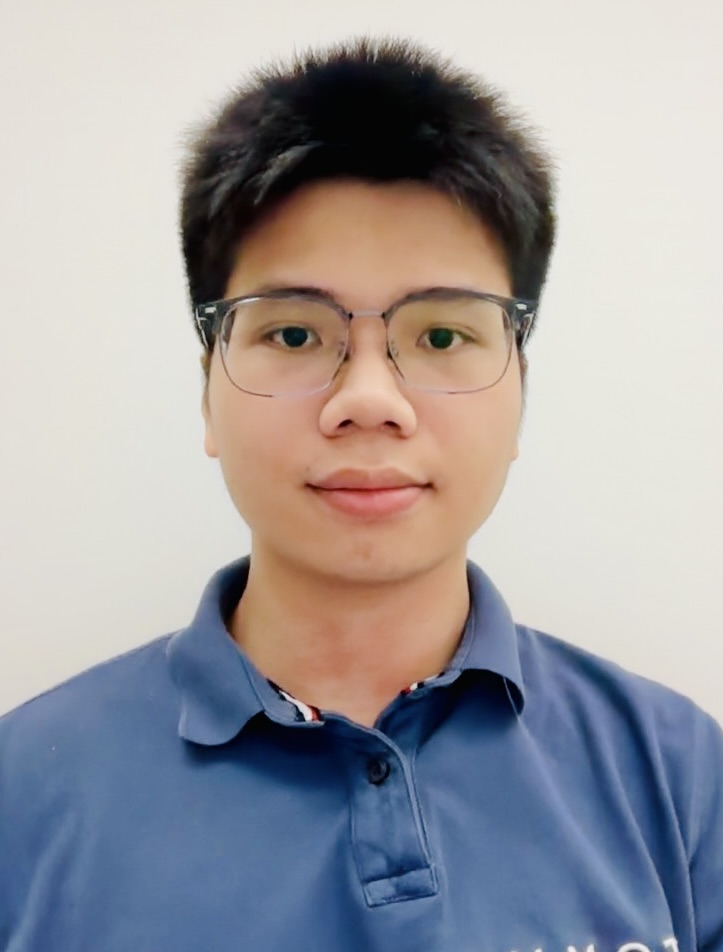}}]{Tue M. Cao}
is currently working toward the Ph.D. degree with the Department of Computer and Information Science and Engineering, University of Florida, Gainesville, FL, USA, as a member of the Adaptive Learning and Optimization Laboratory led by Prof. My T. Thai. His research interests include trustworthy and explainable artificial intelligence, mechanistic interpretability of large language models, and vision–language learning.
\end{IEEEbiography}
\vspace{-5mm}
 
\begin{IEEEbiography}[{\bioPhoto{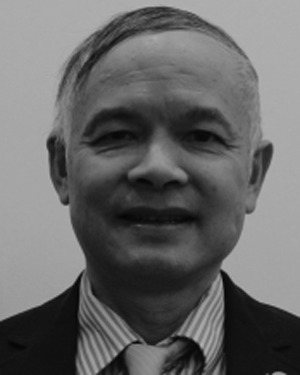}}]{Tien Van Do}
is a Professor with the Department of Networked Systems and Services, Budapest University of Technology and Economics, Budapest, Hungary. He received the Doctor of Science (DSc) title from the Hungarian Academy of Sciences in 2011 and has participated in numerous research projects funded by the European Union. His research interests include performance evaluation of communication networks, queueing theory and stochastic modeling, cloud-native systems and network function virtualization, and machine learning.
\end{IEEEbiography}
 
\begin{IEEEbiography}[{\bioPhoto{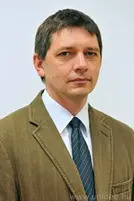}}]{Andr\'as Hajdu}
is a Full Professor with the Faculty of Informatics, University of Debrecen, Debrecen, Hungary, where he heads the Department of Data Science and Visualization. He received the Ph.D. degree from the University of Debrecen in 2003, and was a Post-Doctoral Researcher with the Aristotle University of Thessaloniki, Greece, in 2005--2006. He has been a Full Professor since 2017. His research interests include data science, artificial intelligence, machine and deep learning, medical image analysis, and discrete geometry, with notable contributions to ensemble-based systems for automatic screening of diabetic retinopathy.
\end{IEEEbiography}
\vspace{-5mm}
 
\begin{IEEEbiography}[{\bioPhoto{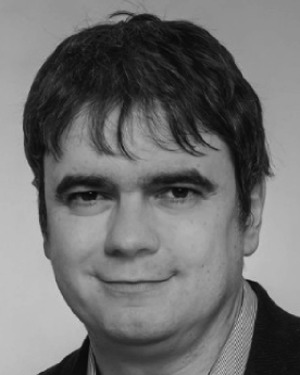}}]{Tam\'as B\'erczes}
(Tam\'as M\'arton B\'erczes) received the Ph.D. degree in informatics from the University of Debrecen, Debrecen, Hungary. He is currently an Associate Professor with the Faculty of Informatics, University of Debrecen. His research interests include queueing theory and retrial queueing systems, performance modeling of computer and communication networks, and the application of machine learning and neural networks to data-driven problems.
\end{IEEEbiography}
\vspace{-5mm}
 
\begin{IEEEbiography}[{\bioPhoto{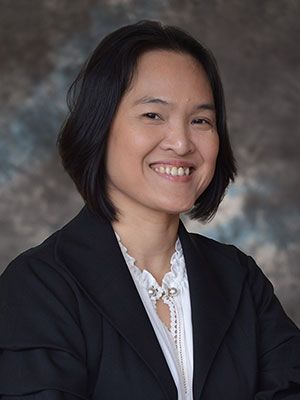}}]{My T. Thai}
(Fellow, IEEE) received the Ph.D. degree in computer science from the University of Minnesota, Twin Cities, MN, USA, in 2005. She is a UF Research Foundation Professor with the Department of Computer and Information Science and Engineering and Associate Director of the Warren B. Nelms Institute for the Connected World, University of Florida, Gainesville, FL, USA. Her research interests include trustworthy AI, billion-scale data mining, optimization, and complex network analysis, with applications to blockchain, social media, cybersecurity, critical networking infrastructure, and healthcare. Her work has resulted in seven books and more than 300 publications. Dr. Thai was a recipient of the Defense Threat Reduction Agency Young Investigator Award in 2009, the NSF CAREER Award in 2010, the UF Research Foundation Professorship in 2016, and several best-paper awards, including the 2023 AAAI Distinguished Paper Award. She serves as Editor-in-Chief of \emph{ACM Computing Surveys} and of the \emph{Journal of Combinatorial Optimization}, and as Founding Editor-in-Chief of \emph{IET Blockchain}.
\end{IEEEbiography}

\vfill


\end{document}